\documentclass{article}
\pdfoutput=1
\usepackage{arxiv}

\usepackage[utf8]{inputenc} 
\usepackage[T1]{fontenc}    
\usepackage{hyperref}       
\usepackage{url}            
\usepackage{booktabs}       
\usepackage{amsfonts}       
\usepackage{microtype}      
\usepackage{xcolor}         
\usepackage{graphicx}
\usepackage{amsmath}
\usepackage{bm}
\title{Formatting Instructions For NeurIPS 2025}

\title{Spiking Neural Network: a low power solution for physical layer authentication}

\author{%
  Jung H. Lee \\
  Pacific Northwest National Laboratory\\
  Seattle, WA \\
  \texttt{jung.lee@pnnl.gov} \\
  \And
  Sujith Vijayan \\
  School of Neuroscience\\
  Virginia Tech\\
  Blacksburg, VA \\
  \texttt{neuron99@vt.edu} \\
}

\begin{document}

\maketitle

\begin{abstract}
Deep learning (DL) is a powerful tool that can solve complex problems, and thus, it seems natural to assume that DL can be used to enhance the security of wireless communication. However, deploying DL models to edge devices in wireless networks is challenging, as they require significant amounts of computing and power resources. Notably, Spiking Neural Networks (SNNs) are known to be efficient in terms of power consumption, meaning they can be an alternative platform for DL models for edge devices. In this study, we ask if SNNs can be used in physical layer authentication. Our evaluation suggests that SNNs can learn unique physical properties (i.e., `fingerprints’) of RF transmitters and use them to identify individual devices. Furthermore, we find that SNNs are also vulnerable to adversarial attacks and that an autoencoder can be used clean out adversarial perturbations to harden SNNs against them.
\end{abstract}

\section{Introduction}

Wireless communication enables ubiquitous hyper-connected networks \cite{IoT}, and as they send signals into open space, security challenges arise. Authentication via public key cryptography (e.g., passwords, tokens or encryption keys) has been used to protect wireless communication, but cryptography-based authentication requires extensive computing power, making it less suitable for edge devices in rapidly growing wireless networks, which are often referred to as Internet of Things (IoT). 

By contrast, Physical Layer Authentication (PLA) uses unique physical properties of Radio Frequency (RF) devices, such as hardware impairments, to identify the devices \cite{PLA1,PLA2} and has been thought to be an effective authentication solution for IoT. Earlier studies \cite{rf-finger, wifidata} showed that deep learning (DL) can automatically detect hardware impairments and other distinct properties of radio hardware devices (e.g., transmitters) and use them to identify individual devices, which is often referred to as ‘RF fingerprinting’. 

However, deploying DL onto edge devices with limited computing power and power supply would be challenging because traditional DL models rely on a massive number of linear and nonlinear operations, which would need advanced hardware and consume substantial amounts of power. Additionally, due to DL models’ well-known vulnerabilities to ‘adversarial’ perturbations \cite{reviewadver1, reviewadver2}, DL models performing as PLA may easily be manipulated by attackers using adversarial perturbations. Then, how do we use DL for PLA? We need PLA systems that will consume small amounts of power and be robust to adversarial attacks. 

As Spiking Neural Networks (SNNs), where neurons communicate with one another via binary signals (“spikes”), consume much less power than traditional DL models do \cite{snn_intro, snn-power}, we ask two questions. First, can SNNs learn to identify RF devices? Second, can we protect SNNs from adversarial attacks? To answer these questions, we train traditional DL models (Artificial Neural Networks, ANNs), convert them to SNNs \cite{ann2snn-1,ann2snn-2, ann2snn-3,qcsf} and test their operation using adversarial inputs. Our empirical evaluation suggests that converted SNNs are highly accurate to identify RF emitters but they can also be vulnerable to adversarial attacks. Here, we propose an effective adversarial defense algorithm that can protect DL models trained to identify RF devices (RF fingerprinting). 

\section{Related Works}
In this study, we 1) test if SNNs can identify individual RF transmitters from their fingerprints (i.e., physical properties), 2) evaluate whether they are vulnerable to adversarial attacks and 3) propose a potential defense algorithm. Below we summarize the earlier works related to our study.

\subsection{RF signals}
In wireless communications, messages are encoded and mapped onto complex numbers, each of which corresponds to a symbol in a constellation diagram. RF transmitters send these symbols via in-phase ($I$) and quadratic ($Q$) phase channels \cite{pysdr}. Notably, the exact waveforms of transmitted signals also depend on unique physical properties of hardware. Even when transmitters and receivers are manufactured through identical processes, they come out with imperfections, adding unique characteristics to transmitted/received RF signals. Communication systems are built to ignore device-specific characteristics, but those in RF signals can serve as fingerprints of RF transmitters, making PLA possible. 

As RF signals encode arbitrary messages, it is necessary to split message-specific and device-specific waveforms to detect RF device fingerprints, but we do note that identical waveforms named `preambles' are added to RF signals to synchronize clocks between transmitters and receivers \cite{pysdr}. As preambles encode predefined messages, it is simpler to extract fingerprints from them. Thus, PLA commonly uses preambles to perform RF fingerprinting, and we can use them to test SNN-based PLA. 

\subsection{Spiking Neural Networks}
Artificial neural networks (ANNs) are loosely inspired by the parallel computing structure of the brain (i.e., biological neural networks) \cite{Hertz, Lecun2015}. The most notable difference between artificial and biological neural networks is computing nodes/units. ANNs’ neurons (i.e., nodes) accept continuous inputs and generate continuous outputs, whereas biological neurons accept and create binary `spikes’; see  \cite{snn_intro} for a reference. That is, spikes mediate information in biological networks, and this spike-based computation is highly energy efficient \cite{snn-power}. SNNs mimic biological networks’ operations by using artificial spiking neurons mimicking biological neurons. With these spiking neurons, SNNs consume much less power than ANNs do.  

However, spikes are discrete and non-differentiable, which makes backpropagation inadequate to training SNNs. Earlier literature proposes two types of alternative algorithms to address this problem. First, Surrogate Gradient Methods (SGM) approximate the activation function of spiking neurons, which are Dirac-delta function, by using steep nonlinear functions such as Heaviside step function \cite{surrogate, surrogate2}. With approximated but differentiable activations, backpropagation can be used to train SNNs. Second, ANNs have been converted to SNNs by transferring ANNs’ weights onto SNNs \cite{ann2snn-1,ann2snn-2,ann2snn-3}. SGM has been used to train SNNs on relatively simple tasks such as recognition of handwritten digits \cite{slayerspikelayererror, surrogate}, whereas converted SNNs can perform more complex tasks more reliably. Thus, we focus on the conversion method in this study. 

Most conversion methods are based on the idea that `Integrate-and-Fire (IF)' neurons can approximate ANNs' ReLU activation functions by using `soft reset' mechanism (see eqs. 1-3); see also \cite{qcsf}.
 \begin{equation}
 \bm{m}^l(t)=\bm{v}^l(t-1)+\bm{W}^{l}\bm{x}^{l-1}(t),
 \end{equation}
 \begin{equation}
 \bm{s}^l(t)=H (\bm{m}^l(t)-\bm{\theta}^l), 
 \end{equation}
 \begin{equation} 
 \bm{v}^{l}(t)=\bm{m}^l(t)-\bm{s}^l(t)\theta^l.   
 \end{equation}
, where $x$ denotes synaptic inputs, and $W$ denotes weight matrix; where $m^l$ and $v^l$ denote membrane potentials before and after a spike generation at t; where $s$ denotes the output spike, and $H$ denotes Heaviside step function; where $\theta$ denotes the threshold for spike generation, and $l$ denotes the layer.

If the weights of ANNs are transferred to SNNs, SNNs’ performance is generally lower than that of the original ANNs. Diehl et al. \cite{weight-normalization} noted that SNNs and ANNs neurons have different activation ranges, which accounts for SNNs’ reduced accuracy, and proposed weight normalization as a solution. Since then, multiple variations of weight normalization have been proposed. Notably, Rueckauer et al. \cite{clipping} proposed that ANN-SNN conversion error can be attributed to quantization errors and that it grows from one layer to another. The membrane potentials $V$ of spiking neurons correspond to ANN neurons’ activation values. When presynaptic spikes arrive, $V$ is increased by a discrete quanta only. That is, membrane potentials are quantized, and consequently, they can only approximate continuous ReLU functions of ANNs. Additional conversion errors were also proposed to be elicited by uneven spikes across time steps and finite ranges of membranes. 

A recent study \cite{qcsf} built upon these findings and further proposed that conversion errors can be minimized if ANNs are trained with activation functions termed ‘Quantization Clip-Floor-Shift (QCFS)’ activation that can approximate behaviors of SNNs’ membrane potentials; see Eq. \ref{qcsf_eq}. 
\begin{equation}
    \label{qcsf_eq}
	\bm{a}^{l} = \widehat{h}(\bm{z}^{l})=\lambda^l~\mathrm{clip} \left(\frac{1}{L}\left \lfloor \frac{\bm{z}^{l}L}{\lambda^l}+ \bm{\varphi} \right \rfloor, 0, 1 \right).
\end{equation}

, where $a^l$ is an activation of IF neurons, and $z$ denotes synaptic inputs; where $L$ is a quantization level, and $\lambda$ determines the maximum activation function of $a^l$; see \cite{qcsf} for more details.

The empirical evaluation suggests that QCFS activation function can minimize the conversion error even when the time step of inference is small \cite{qcsf}. Thus, we adopt QCFS activation function to build and train ANNs to perform RF fingerprinting and convert them into SNNs.
 
\subsection{Earlier SNNs trained on RF fingerprinting}
SNNs have been commonly examined for computer vision \cite{ann2snn-1, surrogate, ann2snn-2}. Recent studies explored SNN-based PLA and used feature extractors. Jiang and Sha \cite{snn-rf2} used SNNs to identify devices in a VHS data exchange system used for global navigation. They used multiple preprocessing steps including spectrograms to train ANNs and convert them to SNNs. Smith et al. \cite{snn-rf} adopted an event detection method and developed acoustic models to detect events in RFs. However, we note that these feature extractors require complex computations, which can negate the very advantage of SNNs. Thus,  we ask if SNNs can learn RF fingerprints from raw RF waveforms after ANN-to-SNN conversion. 

\section{Methods}

\subsection{RF dataset}
We use publicly available WiFi dataset \cite{wifidata}  recorded in a ORBIT testbed, in which 400 nodes (transmitters and receivers) spread over a 20-by-20 grid \cite{orbit}. In the dataset \cite{wifidata}, a center node was always chosen as a receiver and used all other nodes as transmitters. The dataset provides 256 $IQ$ samples collected from 163 transmitters. That is, an individual sample is a 2-dimensional array whose shape is (2,256). The same device was recorded multiple times over 4 days to account for the data drift over time. We note that some devices were recorded for less than 4 days, and the number of signals varied over devices. In this study, we randomly select 100 devices. To minimize the biases from the devices and ensure proper evaluation, we create 10 sets of 100 devices and train an independent model on each dataset. Below, we report the results from 10 models, each of which is trained on a distinct dataset. 

\subsection{Model architecture}
Our fingerprinting model consists of two Residual blocks, each of which has 3 convolutional layers. Table \ref{table1} shows its structure. Following earlier conversion models \cite{qcsf,snn-imagenet-firstinput}, the continuous RF signals are introduced to the first convolutional layer. That is, a single layer of continuous convolutional operation is necessary. This may increase the power consumption, but it can also remove intermediate steps and extra computations necessary for conversions from analog signals to spikes and vice versa. For all our experiments, we use Pytorch, an open source machine learning library \cite{Paszke2017} and use a consumer grade desktop equipped with Core I9 CPU and RTX4090 GPU. Its CPU and GPU have 64 GB and 24 GB memory, respectively. 
\begin{table}[]
\caption{Architecture of our model trained on RF fingerprints. This architecture is inspired by the model used in the earlier study focusing on open-set classification \cite{wifidata}.}\label{table1}
\begin{center}
\begin{tabular}{|llll|l}
\cline{1-4}
Layer                            & {Sub} & {Output Shape}         & {Param \#} &  \\ \cline{1-4}
Residual 1                       &                            &                                             &                                 &  \\ \cline{1-4}
|                                & Conv2d                     & {[}-1, 16, 2, 256{]}                        & 32                              &  \\
|                                & IF                         & {[}-1, 16, 2, 256{]}                        & --                              &  \\
|                                & Conv2d                     & {[}-1, 16, 2, 256{]}                        & 1,552                           &  \\
|                                & BatchNorm2d                & {[}-1, 16, 2, 256{]}                        & 32                              &  \\
|                                & IF                         & {[}-1, 16, 2, 256{]}                        & --                              &  \\
|                                & Conv2d                     & {[}-1, 16, 2, 256{]}                        & 1,552                           &  \\
|                                & BatchNorm2d                & {[}-1, 16, 2, 256{]}                        & 32                              &  \\
|                                & IF                         & {[}-1, 16, 2, 256{]}                        & --                              &  \\
|                                & Conv2d                     & {[}-1, 16, 2, 256{]}                        & { 1,552}    &  \\ \cline{1-4}
{AvgPool2d} & {}    & {{[}-1, 16, 2, 128{]}} & { --}       &  \\ \cline{1-4}
Residual 2                       & {[}-1, 32, 2, 128{]}       & --                                          &                                 &  \\ \cline{1-4}
|                                & Conv2d                     & {[}-1, 32, 2, 128{]}                        & 544                             &  \\
|                                & IF                         & {[}-1, 32, 2, 128{]}                        & --                              &  \\
|                                & Conv2d                     & {[}-1, 32, 2, 128{]}                        & 6,176                           &  \\
|                                & BatchNorm2d                & {[}-1, 32, 2, 128{]}                        & 64                              &  \\
|                                & IF                         & {[}-1, 32, 2, 128{]}                        & --                              &  \\
|                                & Conv2d                     & {[}-1, 32, 2, 128{]}                        & 6,176                           &  \\
|                                & BatchNorm2d                & {[}-1, 32, 2, 128{]}                        & 64                              &  \\
|                                & IF                         & {[}-1, 32, 2, 128{]}                        & --                              &  \\
|                                & Conv2d                     & {[}-1, 32, 2, 128{]}                        & {6,176}    &  \\ \cline{1-4}
{AvgPool2d} & {}    & {} {[}-1, 32, 1, 42{]}  &                                 &  \\ \cline{1-4}
Linear                           & {[}-1, 100{]}              & 134,500                                     &                                 &  \\ \cline{1-4}
\end{tabular}
\end{center}
\end{table}

\subsection{Adversarial attacks and defenses} 
Gradients are generally used to train DL models, but they can also be used to mislead them to incorrect decisions. Since the first attack (the fast gradient sign method) was proposed \cite{goodfellow2015explainingharnessingadversarialexamples}, various types of adversarial attacks have been demonstrated \cite{reviewadver1,reviewadver2}. To craft adversarial inputs (i.e., preambles), we use PGD (projected gradient descent) \cite{pgd}. In each of PGD, the input $x$ is updated by the rule (Eqs. \ref{eq:pgd}). 
\begin{equation}\label{eq:pgd}
x_{t+1} = \Pi_{X}(Z), \textrm{where } Z=x_t + \alpha \cdot sign(\nabla_x J(\theta, x_t, y))
\end{equation}

, where $y$ denotes input and label, $x_0$ and $x_t$ denote original and perturbed input at epoch $t$, $\alpha$ denotes the amplitude of update in each epoch ($\alpha=0.001$ in the experiment), $\theta$ denotes model parameters, $sign$ denotes the sign function, $\nabla_xJ$ denotes the gradient respect to $x$, $\Pi_x$ is a projection function, which restrains the amplitude of perturbation in this study. We use Advertorch \cite{ding2019advertorch}, an open-source adversarial toolbox, to craft 500 adversarial inputs (preambles) using test examples in the dataset. 

Multiple lines of studies have proposed algorithms to make DL models robust to adversarial perturbations. The first line of research proposed adversarial training \cite{Madry2019,zhao2024adversarialtrainingsurvey}, which incorporates adversarial examples with training sets. The adversarial examples are provided with correct labels to teach DL models to predict correct labels of inputs even with adversarial perturbations. It is considered one of the most effective defense algorithms that can make DL models robust to adversarial perturbations. However, adversarial training is known to be attack-specific \cite{bai2021recent, schott2018adversariallyrobustneuralnetwork, LN-NLP}. Due to the variety of adversarial attacks, adding every potential type of attack into training examples would be impossible. Furthermore, it is well documented that adversarial training is quite expensive. 

The second line of research proposed detecting adversarial examples \cite{robustdetectionadversarialexamples, detectingadversarialexamplesnearly}. The proposed algorithms are often inspired by the observation that adversarial perturbations change the patterns of DL features. Consequently, statistical tests (such as clustering) are used to distinguish adversarial inputs from normal ones. The third line of research sought ways to clean out adversarial perturbations \cite{cleansing1,nie2022diffusionmodelsadversarialpurification,lee2023robustevaluationdiffusionbasedadversarial}. If adversarial examples are reconstructed to represent same semantic meanings, we can expect non-semantic adversarial perturbations to be removed during reconstruction. A recent study \cite{diffu_pure} shows that diffusion-based adversarial cleaning is highly effective in removing adversarial perturbations. Diffusion models, however, require a lot of denoising steps with intensive computing.

In our study, we note that preambles have predetermined waveforms and that edge devices do not have the resources to implement complex algorithms such as diffusion-based cleaning. Therefore, we ask 1) if autoencoder can reconstruct preambles to filter out adversarial perturbations and 2) if SNNs can implement autoencoder trained to filter out adversarial perturbations. Our autoencoder is intentionally designed to be simple (Table \ref{table2}), which facilitates its SNN implementation. 

\section{Results}
\begin{figure}
  \centering
  \includegraphics[width=5.5in]{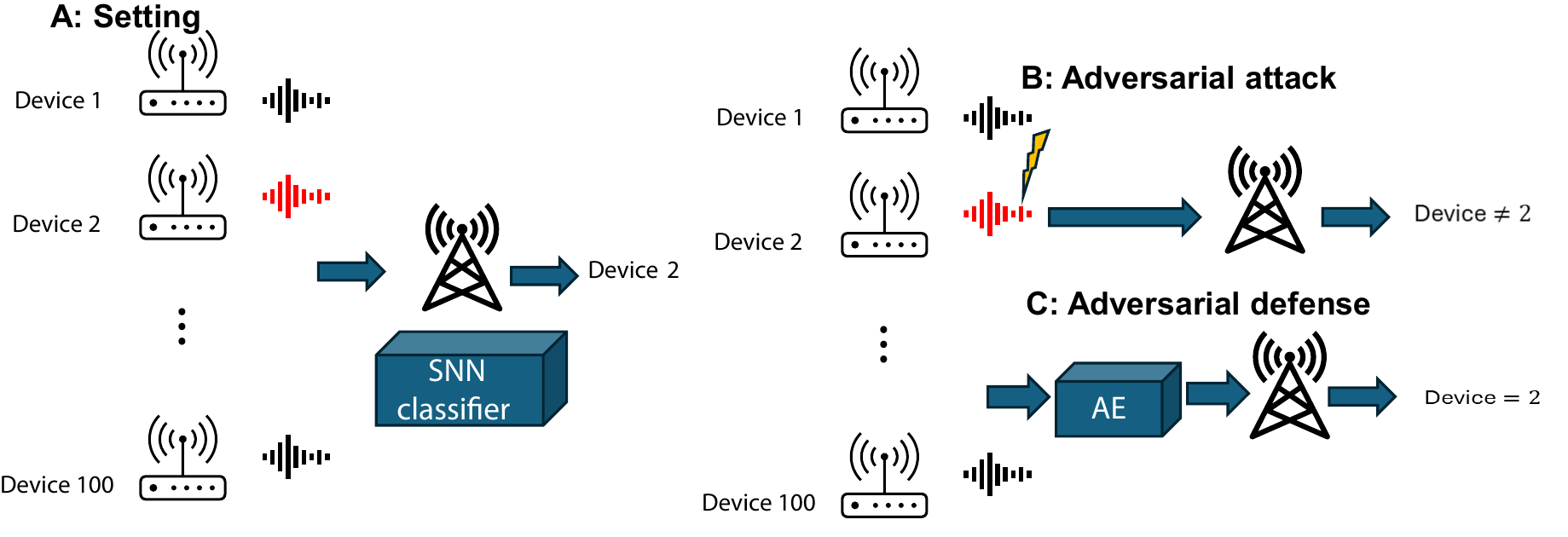}
  \caption{Overview of our experiment. (A), setting of our experiment in Section \ref{exp1}. (B), setting in Section \ref{exp2}. (C), setting in Section \ref{exp3}}\label{diagram}
\end{figure}

We aim to address three questions: 1) Can SNNs be built to recognize RF transmitters from raw RF waveforms? 2) Can an adversarial attack deceive SNNs? 3) Can an autoencoder remove adversarial perturbations? To address these questions, we conduct three sets of experiments. In Section \ref{exp1}, we examine SNNs’ capacity to learn to identify RF device (Fig. \ref{diagram}A). In Section \ref{exp2}, we ask if PGD, a popular adversarial attack, can fool SNNs trained to perform RF device identification (Fig. \ref{diagram}B). In Section \ref{exp3}, we propose a simple defense algorithm against adversarial attacks, which takes advantage of wireless communication systems (Fig. \ref{diagram}C).

\subsection{SNN can learn RF fingerprinting} \label{exp1}
\begin{figure}
  \centering
  \includegraphics[width=5.5in]{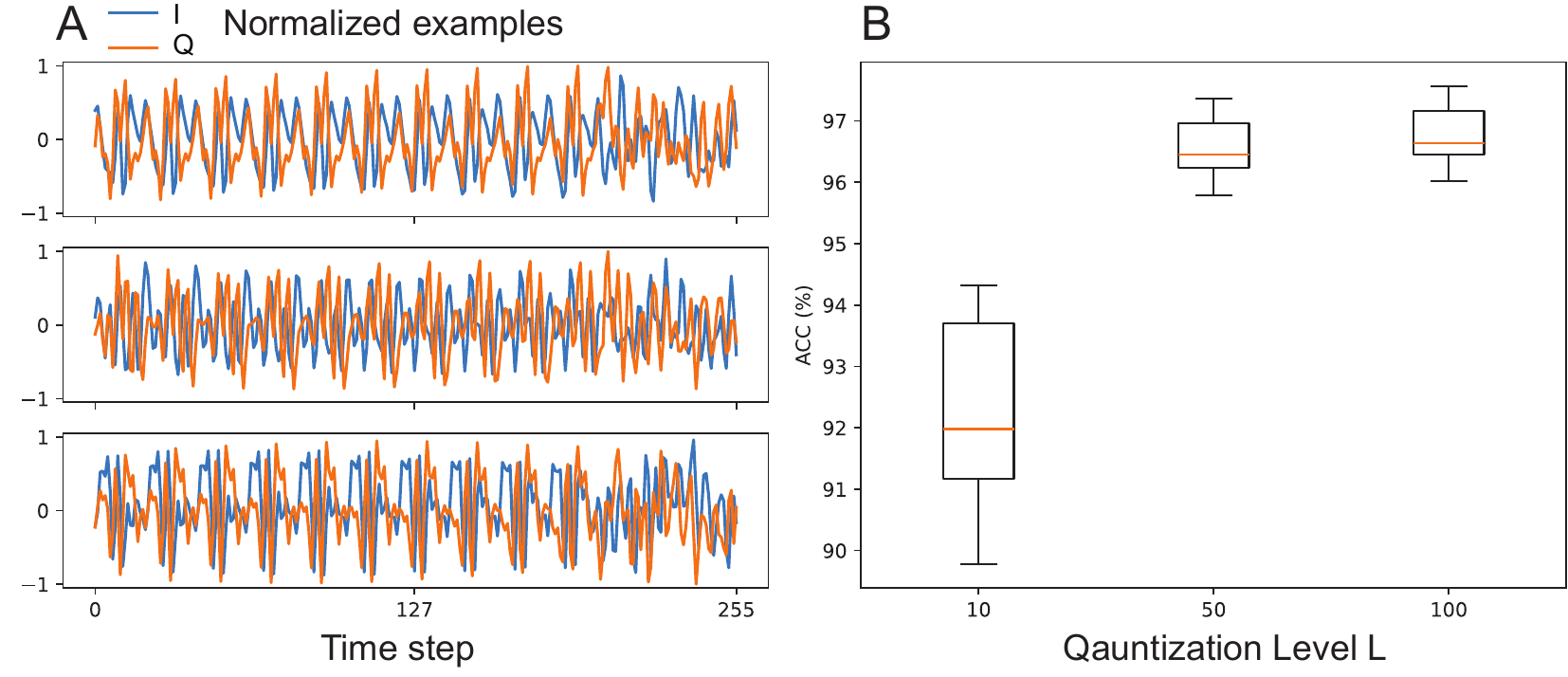}
  \caption{(A), 3 examples of $I$ and $Q$ after normalization. Blue and orange lines represent $I$ and $Q$, respectively. (B), Performance of ANNs trained on 100 RF transmitters. The box plots show $10$ ANNs' accuracy depending on the quantization level ($L$).}\label{fig1}
\end{figure}

We randomly select 100 devices out of 163 and split them into training and test sets. 80\% of the examples are chosen for the training set, and the rest are used as test examples. We aim to minimize preprocessing that can be compute-intensive and thus not be suitable for edge devices. Throughout our study, all RF signals $x$ are normalized to their maximum magnitudes so that RF amplitudes range between -1 and 1 ( $-1\leq x \leq 1$); see Fig. \ref{fig1}A.   

We train ANNs with QCFS activation functions. As exact shapes of QCFS neurons’ activation functions would depend on the quantization level ($L$), we test 3 different levels of quantization $10$, $50$ and $100$. For each quantization, we create $10$ independent training and test sets consisting of randomly chosen 100 devices. As shown in Fig. \ref{fig1}B, ANNs can accurately predict RF devices' identities. 

\begin{figure}
  \centering
  \includegraphics[width=3.5in]{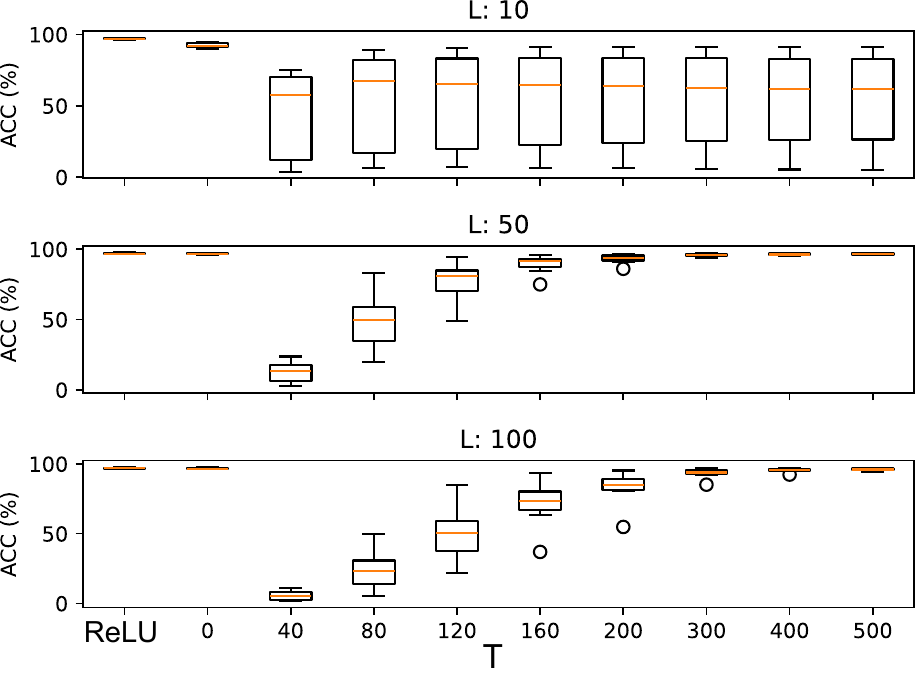}
  \caption{Performance of SNN after conversion. SNNs' accuracy is measured depending on time steps ($T$) used for inference. They are compared with two different ANNs. The first one is ANN with ReLU, and the second one is ANN with QCSF ($T=0$). SNNs' accuracy becomes comparable to ANNs when $T$ is sufficiently high.}\label{fig2}
\end{figure}

Then, we convert ANNs with QCFS activation function to SNNs and evaluate how accurately SNNs can predict the identities of RF devices. Spiking neurons approximately integrate input spikes when soft reset is used (Eqs. 1-3), and their approximations become more accurate when the time step ($T$) increases. Here, we make two germane observations. First, SNNs make more accurate predictions when $L$ is higher (Figs. \ref{fig1}B and \ref{fig2}). With $L$=10, the models’ accuracy cannot go above 55\%. But when $L=50$ or $100$, the accuracy of the models can be comparable to that of ANNs with QCFS (marked by $T=0$) and ReLU. Second, the time step $T$ has strong impact on SNNs’ accuracy, and $T$ should be high to obtain high accuracy (Fig. \ref{fig2}). With $L$=50, the accuracy increases rapidly until $T$=160. These results suggest that SNN-based fingerprinting model can accurately predict RF devices' identities when $L$ is not too low and $T$ is high enough, which is somewhat unexpected, since the original QCFS \cite{qcsf} study showed that low $L$ was sufficient for SNNs to learn complex image classification. 

We do not fully understand the exact reason why our models require moderate $L$ and high $T$ unlike the image classification, but this discrepancy may be explained by the difference between images and RF signals. Specifically, images are encoded using 8 bit integers, and their pixels are spatially organized. By contrast, RF signals are encoded using float variables, and the difference between RF fingerprints of the devices (e.g., classes) is very small. Based on these facts, we speculate that high-precision computation is necessary for processing RF signals, which requires moderate $L$ and high $T$.

\subsection{SNNs are also vulnerable to adversarial attacks}\label{exp2}
The results above suggest that SNNs can learn RF devices’ distinct characteristics and use them to identify RF devices. However, PLA systems, a gateway for sensitive information in wireless networks, should be resilient against malicious activities or tempering. Noting the fact that DL models are vulnerable to adversarial perturbations, we test if adversarial attacks can disrupt SNNs’ abilities to identify RF devices. Specifically, we use PGD, a common gradient-based adversarial attacks \cite{pgd}, to craft 500 adversarial RF signals for ANNs with QCFS activation. We restrict the maximum perturbation strength ($\epsilon$). As signal $x$ is normalized between -1 and 1, $\epsilon=0.1$ means that the magnitude of perturbations can be 10 \% of the signal strength. 

\begin{figure}
  \centering
  \includegraphics[width=3.5in]{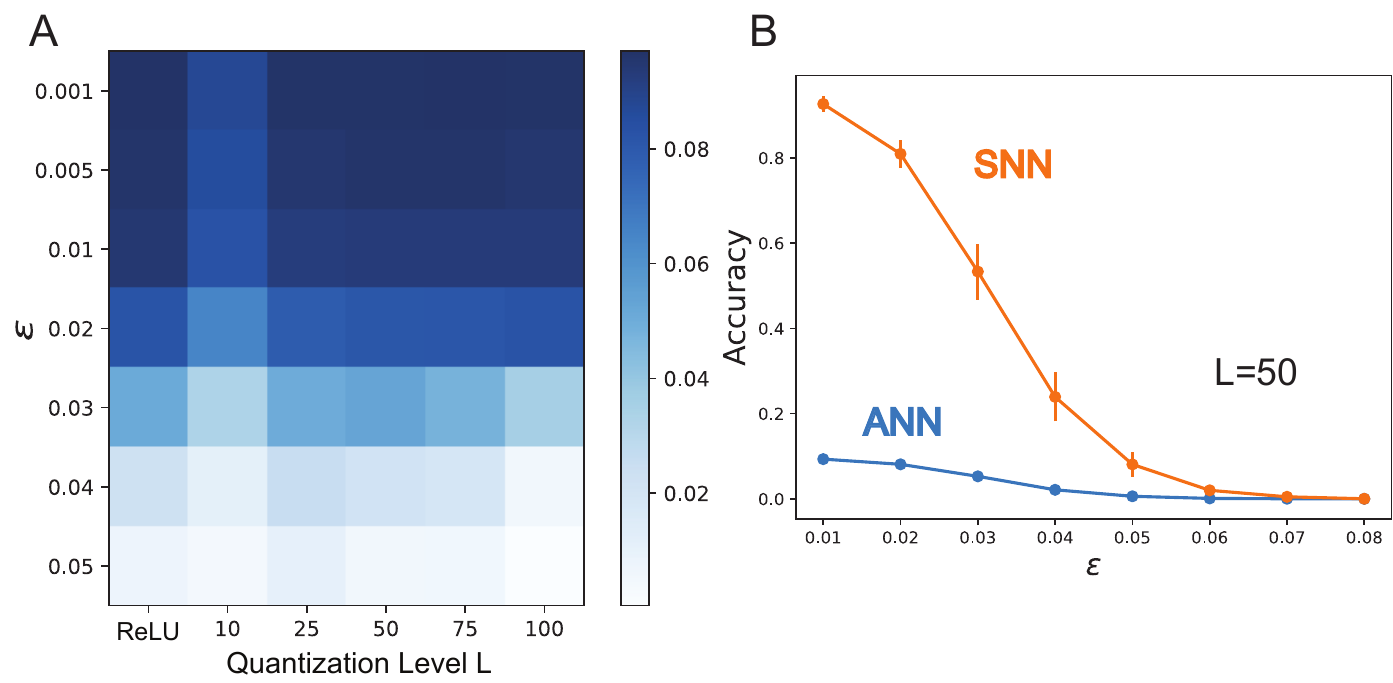}
  \caption{(A), Accuracy of ANNs’ predictions on adversarial inputs depending on $\epsilon$ and $L$. The color indicates the accuracy averaged over 10 models.  All adversarial signals are crafted for specific models (i.e., white-box attack). When we test the models’ QCFS activation function, we use $L=10,25, 50, 75, 100$.  The column, marked by $ReLU$, indicates the model with ReLU activation function. For both models,  accuracy is around 0.1 or below it. (B), Comparison between ANNs and SNNs with $L=50$. The accuracy levels of ANNs and SNNs are displayed in blue and orange. }\label{fig3}
 
\end{figure}

In the experiment, we craft adversarial examples by varying $\epsilon$ and $L$. Fig. \ref{fig3}A shows the accuracy of ANNs with both ReLU and QCFS activation function (see columns with $L>0$) on adversarial inputs, suggesting that these adversarial signals can effectively reduce the accuracy of ANNs regardless of quantization levels. Then, we test whether SNNs, converted from corresponding ANNs, could also be affected by the same adversarial inputs. As shown in Fig. \ref{fig3}B, SNN appear more robust to adversarial examples than ANNs, when $\epsilon \approx 0.01$, but the accuracy gap rapidly decreases, as $\epsilon$ increases.

\subsection{Autoencoder can remove adversarial perturbations from preambles}\label{exp3}
As adversarial signals can be transferred from an ANN to a SNN counterpart, it is essential that we strengthen SNNs to be robust to adversarial defense algorithms. We note that high computing load or power consumption should not be required for adversarial defense of SNN-based PLA systems. Then, how do we effectively and economically protect SNN-based PLA from adversarial attacks? Based on the fact that PLA uses RF fingerprints in preambles (identical messages across devices), we hypothesize that we may effectively remove adversarial perturbations from preambles by reconstructing preambles. 

\begin{figure}
  \centering
  \includegraphics[width=3.5in]{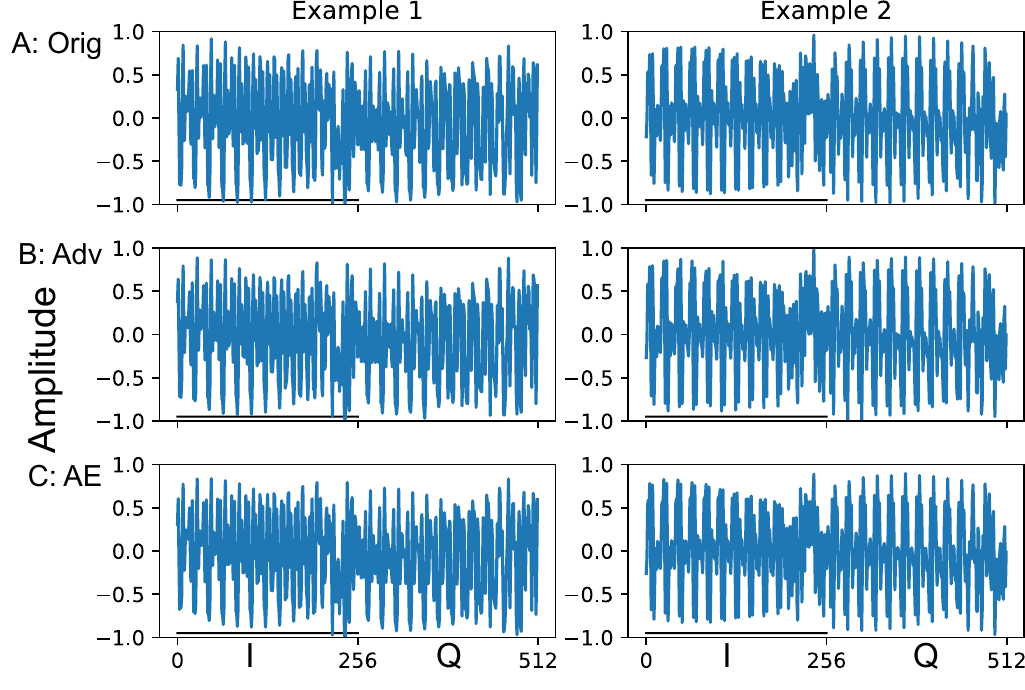}
  \caption{Examples signals. (A), 2 examples of clean signals. (B), 2 examples of adversarial examples. (C), adversarial signals cleaned out by AE. The black horizontal line denotes the $I$ samples, and the rest denote $Q$ samples.}\label{example}
  \end{figure}

Diffusion-based cleansing may be one of the best reconstruction-based adversarial defenses \cite{diffu_pure}, but diffusion models are not suitable for edge devices because they need intense computations. Thus, we test if an autoencoder (AE) can be trained to reconstruct preambles and clean adversarial perturbations. The architecture of an autoencoder is illustrated in Table \ref{table2}. It should be noted that our AE reconstructs $I$ and $Q$ samples together by flattening 2 dimensional input vectors consisting of $I$ and $Q$ symbols into 1 dimensional input vectors (Fig. \ref{example}). Consequently, the inputs for AE are 512 dimensional vectors and go through an encoder (3 fully connected layers) to create latent variables, which in turn go through a decoder (3 fully connected layers) to generate reconstructed inputs. Fig. \ref{example} shows examples of 1) clean, 2) adversarial and 3) adversarial cleaned out by AE. 

\begin{figure}
  \centering
  \includegraphics[width=3.5in]{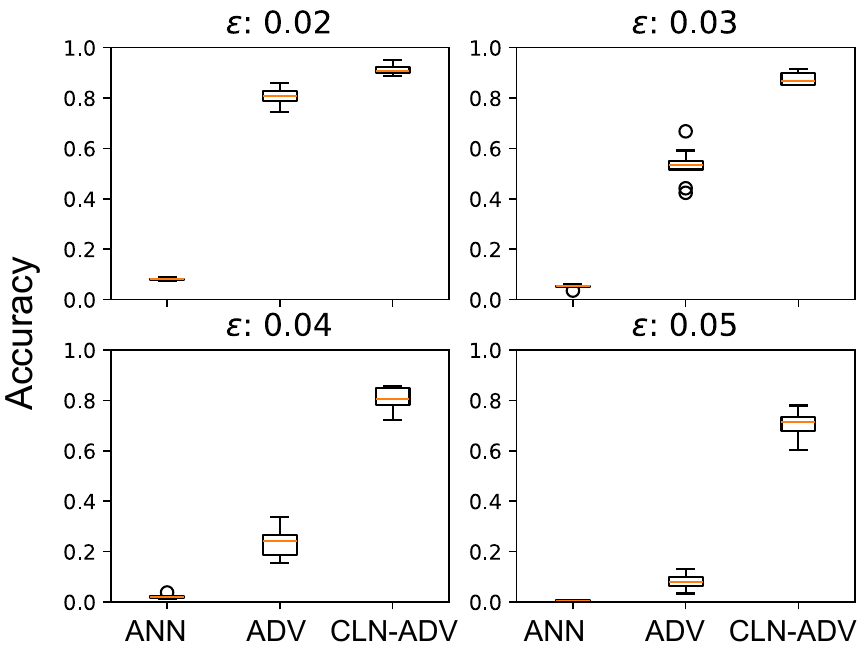}
  \caption{Accuracy of SNN with and without autoencoder. The box plots compare the accuracy of ANNs on adversarial signals (ANN), the accuracy of SNNs on adversarial signal without AE (ADV) and the accuracy of SNNs with AE (CLN-ADV). We display the accuracies for 4 different $\epsilon$.}\label{fig4}
  
\end{figure}

We train AE to reconstruct RF signals from training set. In the experiment, we train 10 AEs using 10 training sets mentioned above. After training AEs, we covert them to SNNs, which we use to clean the preambles. That is, AEs are also implemented using SNNs. We accumulate outputs of SNN-based AE over 400 time steps (i,e., $T=400$) to reconstruct preambles. The reconstructed preambles are used to do fingerprinting. Fig. \ref{fig4} shows the accuracy of SNN-based fingerprinting models with an AE. 

\begin{figure}
  \centering
  \includegraphics[width=2.5in]{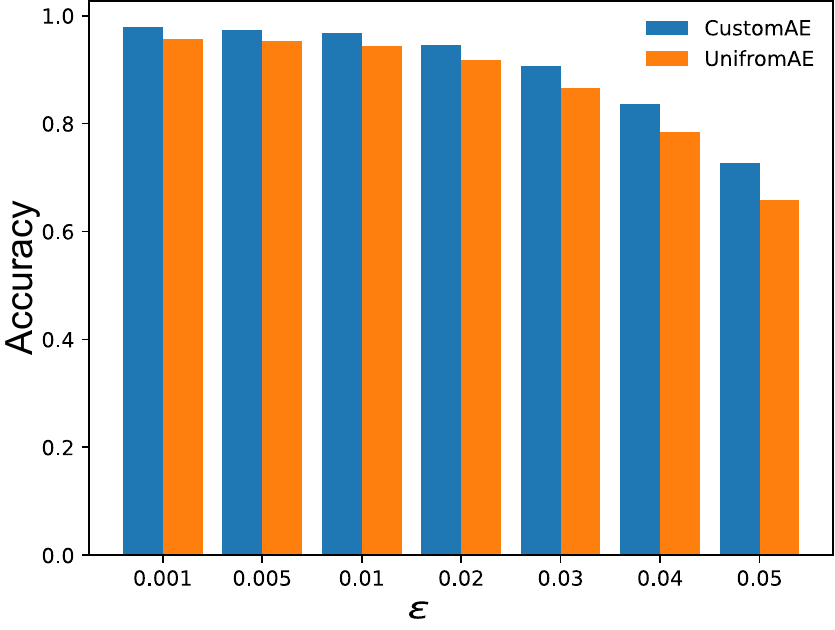}
  \caption{Transferability of AE. The orange bars denote the accuracy of SNNs, when a universal AE cleans inputs, whereas the blue bars denote the accuracy of SNNs, when AE, individually trained on 100 devices, cleans inputs.  We display the mean value of 10 models' accuracies.}
  \label{fig5}
\end{figure}

For all 4 cases ($\epsilon=0.02-0.05$), SNNs' predictions (marked by `ADV' in the figure) are more accurate than those of ANNs (marked by `ANN' in the figure), but as $\epsilon$ increases, the accuracy of SNNs' predictions decreases significantly. Importantly, the accuracy of SNNs' predictions on preamble reconstructed by AE is $\approx 80\%$ or higher.  Finally, we ask if AE should be trained for a specific set of devices. To this end, we train a single AE on RF signals from 50 devices randomly chosen and use this universal AE to reconstruct preambles for all 10 SNN-based fingerprinting models, which are trained on different sets of RF emitters. Fig. \ref{fig5} shows the accuracy (orange bars in the figure) of SNNs' predictions on preambles reconstructed by this universal AE . For comparison, we also report the accuracy (blue bars in the figure) of SNNs' predictions on preambles reconstructed by AEs that are trained on the same $100$ devices. As shown in the figure, they are comparable, suggesting that AE can be transferred across devices. 

\section{Conclusion}
In this study, we propose a SNN-based PLA system for WiFi networks and evaluate its accuracy. Our results raise the possibility that SNNs can be used to build a pipeline for PLA, which is robust to adversarial attacks. Our work is different from the earlier studies that tested SNNs for RF fingerprinting in three regards. First, our model works on raw RF signals instead of features discovered by preprocessing. Second, we evaluate the vulnerabilities of SNNs trained for PLA to adversarial attacks. Third, we propose an effective way to defend SNNs from adversarial attacks. 

\subsection{Limitations}
Our study utilizes RF signals collected from Orbit testbed. This database does contain real RF signals, but the data is collected in an controlled environment. In the real world, multiple deflectors would exist between transmitters and receivers, which would create complex multi-channel effects and interferences between them \cite{pysdr}. Consequently, PLA would deal with much more noisy signals. We plan to evaluate the performance of SNN-based PLA with more complex channel effects in the future. 

\subsection{Broader Impacts}
Sensitive information is constantly shared via wireless networks, and the security of wireless networks is increasingly important these days. As our study indicates that SNNs can enhance the security of wireless networks and present an effective defense against adversarial attacks, we believe more studies on SNN-based security solutions will be encouraged as a result.  

\begin{table}[]
\caption{Architecture of autoencoder trained to reconstruct RF signals (preambles).}\label{table2}
\begin{center}
\begin{tabular}{|llll|}
\hline
Layer   & Sub-layer & {Output Shape}     & {Param \#} \\ \hline
Encoder &           & --                                      &                                 \\ \hline
|       & Linear    & {[}-1, 1, 256{]}                        & 131,328                         \\
|       & IF        & {[}-1, 1, 256{]}                        & --                              \\
|       & Linear    & {[}-1, 1, 256{]}                        & 65,792                          \\
|       & IF        & {[}-1, 1, 256{]}                        & --                              \\
|       & Linear    & {[}-1, 1, 128{]}                        & 32,896                          \\ \hline
Decoder &           & {{[}-1, 1, 512{]}} &                                 \\ \hline
|       & Linear    & {[}-1, 1, 256{]}                        & 33,024                          \\
|       & IF        & {[}-1, 1, 256{]}                        & --                              \\
|       & Linear    & {[}-1, 1, 256{]}                        & 65,792                          \\
|       & IF        & {[}-1, 1, 256{]}                        & --                              \\
|       & Linear    & {[}-1, 1, 512{]}                        & 131,584                         \\ \hline
\end{tabular}
\end{center}
\end{table}

\bibliography{ref}

\begin{thebibliography}{10}

\bibitem{IoT}
S.~Kumar, P.~Tiwari, and M~Zymbler.
\newblock Internet of things is a revolutionary approach for future technology
  enhancement: a review.
\newblock {\em J Big Data}, 2019.

\bibitem{PLA1}
Junqing Zhang, Sekhar Rajendran, Zhi Sun, Roger Woods, and Lajos Hanzo.
\newblock Physical layer security for the internet of things: Authentication
  and key generation.
\newblock {\em IEEE Wireless Communications}, 26(5):92--98, 2019.

\bibitem{PLA2}
Paul~L. Yu, John~S. Baras, and Brian~M. Sadler.
\newblock Physical-layer authentication.
\newblock {\em IEEE Transactions on Information Forensics and Security},
  3(1):38--51, 2008.

\bibitem{rf-finger}
Tong Jian, Bruno~Costa Rendon, Emmanuel Ojuba, Nasim Soltani, Zifeng Wang,
  Kunal Sankhe, Andrey Gritsenko, Jennifer Dy, Kaushik Chowdhury, and Stratis
  Ioannidis.
\newblock Deep learning for rf fingerprinting: A massive experimental study.
\newblock {\em IEEE Internet of Things Magazine}, 3(1):50--57, 2020.

\bibitem{wifidata}
Samer Hanna, Samurdhi Karunaratne, and Danijela Cabric.
\newblock Open set wireless transmitter authorization: Deep learning approaches
  and dataset considerations.
\newblock {\em IEEE Transactions on Cognitive Communications and Networking},
  7(1):59--72, 2021.

\bibitem{reviewadver1}
Anirban Chakraborty, Manaar Alam, Vishal Dey, Anupam Chattopadhyay, and Debdeep
  Mukhopadhyay.
\newblock Adversarial attacks and defences: A survey, 2018.

\bibitem{reviewadver2}
Xiaoyong Yuan, Pan He, Qile Zhu, and Xiaolin Li.
\newblock Adversarial examples: Attacks and defenses for deep learning.
\newblock {\em IEEE Transactions on Neural Networks and Learning Systems},
  30(9):2805--2824, 2019.

\bibitem{snn_intro}
Wulfram Gerstner, Werner~M. Kistler, Richard Naud, and Liam Paninski.
\newblock {\em Neuronal Dynamics: From Single Neurons to Networks and Models of
  Cognition}.
\newblock Cambridge University Press, USA, 2014.

\bibitem{snn-power}
Xinyu Shi, Jianhao Ding, Zecheng Hao, and Zhaofei Yu.
\newblock Towards energy efficient spiking neural networks: An unstructured
  pruning framework.
\newblock In {\em The Twelfth International Conference on Learning
  Representations}, 2024.

\bibitem{ann2snn-1}
Bing Han and Kaushik Roy.
\newblock Deep spiking neural network: Energy efficiency through time based
  coding.
\newblock In Andrea Vedaldi, Horst Bischof, Thomas Brox, and Jan-Michael Frahm,
  editors, {\em Computer Vision -- ECCV 2020}, pages 388--404, Cham, 2020.
  Springer International Publishing.

\bibitem{ann2snn-2}
Shikuang Deng and Shi Gu.
\newblock Optimal conversion of conventional artificial neural networks to
  spiking neural networks, 2021.

\bibitem{ann2snn-3}
Abhronil Sengupta, Yuting Ye, Robert Wang, Chiao Liu, and Kaushik Roy.
\newblock Going deeper in spiking neural networks: Vgg and residual
  architectures.
\newblock {\em Frontiers in Neuroscience}, Volume 13 - 2019, 2019.

\bibitem{qcsf}
Tong Bu, Wei Fang, Jianhao Ding, PengLin Dai, Zhaofei Yu, and Tiejun Huang.
\newblock Optimal ann-snn conversion for high-accuracy and ultra-low-latency
  spiking neural networks, 2023.

\bibitem{pysdr}
Marc Lichtman.
\newblock Pysdr: A guide to sdr and dsp using python.
\newblock \url{https://https://github.com/777arc/PySDR}, 2025.

\bibitem{Hertz}
John Hertz, Richard~G. Palmer, and Anders~S. Krogh.
\newblock {\em Introduction to the Theory of Neural Computation}.
\newblock Perseus Publishing, 1st edition, 1991.

\bibitem{Lecun2015}
Yann Lecun, Yoshua Bengio, and Geoffrey Hinton.
\newblock {Deep learning}.
\newblock {\em Nature}, 521(7553):436--444, 2015.

\bibitem{surrogate}
Emre~O. Neftci, Hesham Mostafa, and Friedemann Zenke.
\newblock Surrogate gradient learning in spiking neural networks, 2019.

\bibitem{surrogate2}
Chenlin Zhou, Han Zhang, Liutao Yu, Yumin Ye, Zhaokun Zhou, Liwei Huang,
  Zhengyu Ma, Xiaopeng Fan, Huihui Zhou, and Yonghong Tian.
\newblock Direct training high-performance deep spiking neural networks: a
  review of theories and methods.
\newblock {\em Frontiers in Neuroscience}, Volume 18 - 2024, 2024.

\bibitem{slayerspikelayererror}
Sumit~Bam Shrestha and Garrick Orchard.
\newblock Slayer: Spike layer error reassignment in time, 2018.

\bibitem{weight-normalization}
Peter~U. Diehl, Daniel Neil, Jonathan Binas, Matthew Cook, Shih-Chii Liu, and
  Michael Pfeiffer.
\newblock Fast-classifying, high-accuracy spiking deep networks through weight
  and threshold balancing.
\newblock In {\em 2015 International Joint Conference on Neural Networks
  (IJCNN)}, pages 1--8, 2015.

\bibitem{clipping}
Bodo Rueckauer, Iulia-Alexandra Lungu, Yuhuang Hu, and Michael Pfeiffer.
\newblock Theory and tools for the conversion of analog to spiking
  convolutional neural networks, 2016.

\bibitem{snn-rf2}
Qi~Jiang and Jin Sha.
\newblock The use of snn for ultralow-power rf fingerprinting identification
  with attention mechanisms in vdes-sat.
\newblock {\em IEEE Internet of Things Journal}, 10(17):15594--15603, 2023.

\bibitem{snn-rf}
Michael Smith, Michael~A Temple, and James Dean.
\newblock Development of a neuromorphic-friendly spiking neural network for rf
  event-based classification, 2015.

\bibitem{orbit}
D.~Raychaudhuri, I.~Seskar, M.~Ott, S.~Ganu, K.~Ramachandran, H.~Kremo,
  R.~Siracusa, H.~Liu, and M.~Singh.
\newblock Overview of the orbit radio grid testbed for evaluation of
  next-generation wireless network protocols.
\newblock In {\em IEEE Wireless Communications and Networking Conference,
  2005}, volume~3, pages 1664--1669 Vol. 3, 2005.

\bibitem{snn-imagenet-firstinput}
Nitin Rathi and Kaushik Roy.
\newblock Diet-snn: A low-latency spiking neural network with direct input
  encoding and leakage and threshold optimization.
\newblock {\em IEEE Transactions on Neural Networks and Learning Systems},
  34(6):3174--3182, 2023.

\bibitem{Paszke2017}
Adam Paszke, Sam Gross, Soumith Chintala, Edward {Chanan, Gregory Yang},
  Zachary DeVito, Alban {Lin, Zeming Desmaison}, Luca Antiga, and Adam Lerer.
\newblock {Automatic differentiation in PyTorch}.
\newblock In {\em NIPS Autodiff Workshop}, 2017.

\bibitem{goodfellow2015explainingharnessingadversarialexamples}
Ian~J. Goodfellow, Jonathon Shlens, and Christian Szegedy.
\newblock Explaining and harnessing adversarial examples, 2015.

\bibitem{pgd}
Aleksander Madry, Aleksandar Makelov, Ludwig Schmidt, Dimitris Tsipras, and
  Adrian Vladu.
\newblock Towards deep learning models resistant to adversarial attacks, 2017.

\bibitem{ding2019advertorch}
Gavin~Weiguang Ding, Luyu Wang, and Xiaomeng Jin.
\newblock {AdverTorch} v0.1: An adversarial robustness toolbox based on
  pytorch.
\newblock {\em arXiv preprint arXiv:1902.07623}, 2019.

\bibitem{Madry2019}
Aleksander Madry, Aleksandar Makelov, Ludwig Schmidt, Dimitris Tsipras, and
  Adrian Vladu.
\newblock {Towards Deep Learning Models Resistant to Adversarial Attacks}.
\newblock In {\em NeurIPS}, page 1706.06083, 2019.

\bibitem{zhao2024adversarialtrainingsurvey}
Mengnan Zhao, Lihe Zhang, Jingwen Ye, Huchuan Lu, Baocai Yin, and Xinchao Wang.
\newblock Adversarial training: A survey, 2024.

\bibitem{bai2021recent}
Tao Bai, Jinqi Luo, Jun Zhao, Bihan Wen, and Qian Wang.
\newblock Recent advances in adversarial training for adversarial robustness,
  2021.

\bibitem{schott2018adversariallyrobustneuralnetwork}
Lukas Schott, Jonas Rauber, Matthias Bethge, and Wieland Brendel.
\newblock Towards the first adversarially robust neural network model on mnist,
  2018.

\bibitem{LN-NLP}
Joel Dapello, Tiago Marques, Martin Schrimpf, Franziska Geiger, David Cox, and
  James~J DiCarlo.
\newblock Simulating a primary visual cortex at the front of cnns improves
  robustness to image perturbations.
\newblock In H.~Larochelle, M.~Ranzato, R.~Hadsell, M.F. Balcan, and H.~Lin,
  editors, {\em Advances in Neural Information Processing Systems}, volume~33,
  pages 13073--13087. Curran Associates, Inc., 2020.

\bibitem{robustdetectionadversarialexamples}
Tianyu Pang, Chao Du, Yinpeng Dong, and Jun Zhu.
\newblock Towards robust detection of adversarial examples, 2018.

\bibitem{detectingadversarialexamplesnearly}
Florian Tramèr.
\newblock Detecting adversarial examples is (nearly) as hard as classifying
  them, 2022.

\bibitem{cleansing1}
Yi-Hsuan Wu, Chia-Hung Yuan, and Shan-Hung Wu.
\newblock Adversarial robustness via runtime masking and cleansing.
\newblock In Hal~Daumé III and Aarti Singh, editors, {\em Proceedings of the
  37th International Conference on Machine Learning}, volume 119 of {\em
  Proceedings of Machine Learning Research}, pages 10399--10409. PMLR, 13--18
  Jul 2020.

\bibitem{nie2022diffusionmodelsadversarialpurification}
Weili Nie, Brandon Guo, Yujia Huang, Chaowei Xiao, Arash Vahdat, and Anima
  Anandkumar.
\newblock Diffusion models for adversarial purification, 2022.

\bibitem{lee2023robustevaluationdiffusionbasedadversarial}
Minjong Lee and Dongwoo Kim.
\newblock Robust evaluation of diffusion-based adversarial purification, 2023.

\bibitem{diffu_pure}
Weili Nie, Brandon Guo, Yujia Huang, Chaowei Xiao, Arash Vahdat, and Anima
  Anandkumar.
\newblock Diffusion models for adversarial purification, 2022.

\end{thebibliography}

\bibliographystyle{unsrt}


\end{document}